\documentclass{article}

\usepackage[preprint]{neurips_2023}

\usepackage[utf8]{inputenc} 
\usepackage[T1]{fontenc}    
\usepackage{hyperref}       
\usepackage{url}            
\usepackage{booktabs}       
\usepackage{amsfonts}       
\usepackage{nicefrac}       
\usepackage{microtype}      
\usepackage[table]{xcolor}        
\usepackage{wrapfig}
\usepackage{caption}
\usepackage{subcaption}
\usepackage{graphicx}
\usepackage{amsmath}
\usepackage{amssymb}

\definecolor{LightCyan}{rgb}{0.9,0.96,1}

\title{HMSN: Hyperbolic Self-Supervised Learning by Clustering with Ideal Prototypes}

\author{%
   Aiden Durrant \qquad Georgios Leontidis\\
   University of Aberdeen\\
   Department of Computing Science \& Interdisciplinary Centre for Data and AI\\Aberdeen, United Kingdom \\
   {\tt \{a.durrant.20, georgios.leontidis\}@abdn.ac.uk} \\
}

\begin{document}

\maketitle

\begin{abstract}
 Hyperbolic manifolds for visual representation learning allow for effective learning of semantic class hierarchies by naturally embedding tree-like structures with low distortion within a low-dimensional representation space. The highly separable semantic class hierarchies produced by hyperbolic learning have shown to be powerful in low-shot tasks, however, their application in self-supervised learning is yet to be explored fully. In this work, we explore the use of hyperbolic representation space for self-supervised representation learning for prototype-based clustering approaches. First, we extend the Masked Siamese Networks to operate on the Poincar\'e ball model of hyperbolic space, secondly, we place prototypes on the ideal boundary of the Poincar\'e ball. Unlike previous methods we project to the hyperbolic space at the output of the encoder network and utilise a hyperbolic projection head to ensure that the representations used for downstream tasks remain hyperbolic. Empirically we demonstrate the ability of these methods to perform comparatively to Euclidean methods in lower dimensions for linear evaluation tasks, whilst showing improvements in extreme few-shot learning tasks.
 
\end{abstract}

\section{Introduction}
\label{sec:intro}
Self-supervised representation learning for natural images has continued to make vast progress in past years \citep{chen2020simple, he2020momentum, tian2020contrastive, caron2021emerging, bardes2021vicreg, assran2023self}, quickly approaching, and in cases with significant data preprocessing surpassing supervised learning performance \citep{tomasev2022pushing}. The advantage of self-supervision lies in the ability to leverage the large quantities of data that exist in the world without human annotations to learn high-quality representations. However, most established self-supervised visual learning methods typically project representations on Euclidean or Hyperspherical manifolds, and in some cases disregarding the underlying hyperbolic structure of the data.


When attempting to capture hyperbolic data structures, zero and positive curvature spaces exhibit some inherent implications as opposed to negative curvature hyperbolic space, most notably the inability to embed hierarchical semantic relationships between points in space, a well-established principle of learning good representations \citep{bengio2013representation}. Although there has been debate to what extent natural images exhibit underlying hyperbolic structure of semantics, recent works have demonstrated via empirical metrics the presence of latent hierarchical tree-like structures in standard computer vision datasets \citep{khrulkov2020hyperbolic, ermolov2022hyperbolic}, and subsequently shown the capabilities of hyperbolic representations to excel in these settings \citep{yan2021unsupervised, khrulkov2020hyperbolic, ermolov2022hyperbolic}. Many of these advancements in hyperbolic learning have been seen in metric and prototype learning settings, specifically for few-shot learning, where the highly separable semantic hierarchies lead to better-performing few-shot classifiers \citep{ghadimiatigh2022hyperbolic}. Self-supervision via prototype learning has also demonstrated state-of-the-art performance in few-shot and low-shot tasks \citep{assran2022masked}, therefore we leverage the hierarchical learning capabilities of hyperbolic prototype learning in the self-supervised setting for improved few-shot learning.

In this work, we propose the use of hyperbolic representation spaces in Self-Supervised Learning (SSL) to more appropriately embed the natural semantic class hierarchies presented in the data. We first demonstrate the capability of hyperbolic learning on a leading low-shot learning method of Masked Siamese Networks (MSNs) \citep{assran2022masked}. Here we project the output Euclidean representation onto the Poincar\'e ball and use the Poincar\'e distance in gyrovector space in place of the cosine similarity in the probability computation of the codes. We empirically show that such a conversion to hyperbolic space can lead to an improvement in representation quality for few-shot downstream tasks. Importantly and unlike previous methods \citep{ge2022hyperbolic, yue2023hyperbolic}, we propose the use of fully hyperbolic projection networks, projecting the output of the encoder to hyperbolic space to ensure the hyperbolicity we aim to learn in the representations is utilised in downstream tasks.

In addition, we propose a new self-supervised method based on MSNs that leverages the advancements in hyperbolic prototype learning \citep{ghadimi2021hyperbolic} where instead of continually learning prototypes, we place the prototypes on the ideal boundary of the Poincar\'e ball of hyperbolic space. We train our network to produce good hyperbolic representations through a new loss function based on the Busemann distance metric \citep{ghadimi2021hyperbolic}. We empirically demonstrate improvements over the Euclidean baseline and our hyperbolic conversion on few-shot and extreme low-shot learning tasks. Furthermore, we show that our hyperbolic methods are competitive with other Euclidean methods through standard self-supervised linear evaluation and transfer learning benchmarks.

\begin{figure}[t]
    \centering
    \begin{subfigure}[b]{0.4\textwidth}
         \centering
         \includegraphics[height=12.5em]{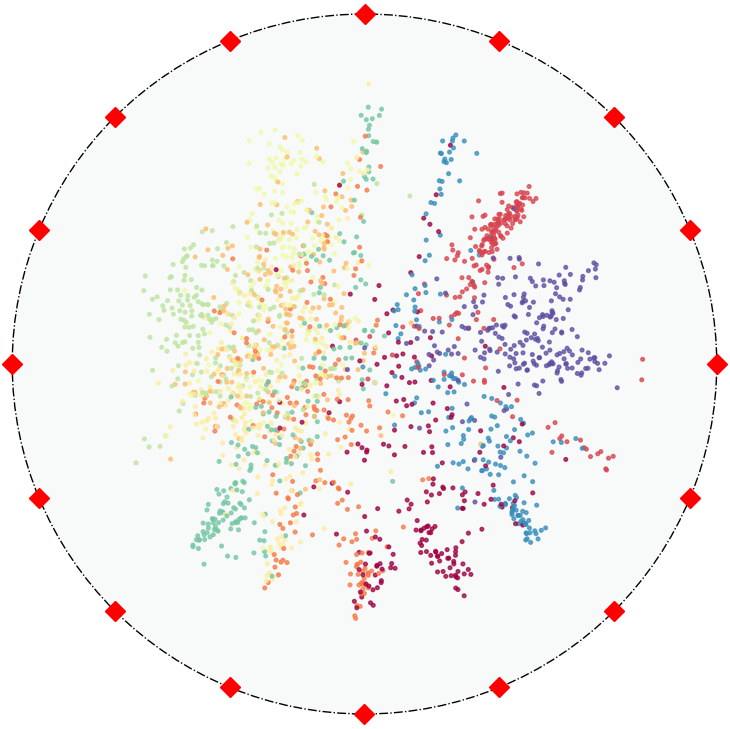}
            \caption{HMSN-IP}
         \label{fig:hmsn-1p_2d}
    \end{subfigure}
    \begin{subfigure}[b]{0.59\textwidth}
         \centering
          \includegraphics[height=13em]{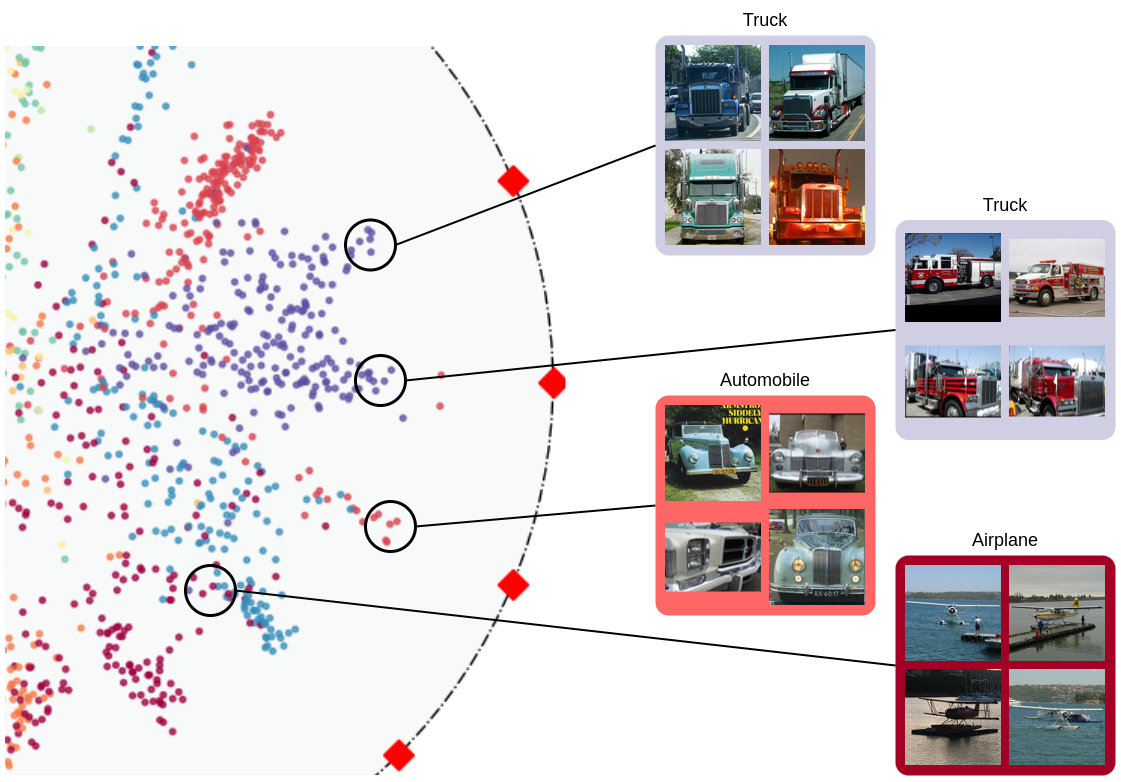}
        \caption{Hierarchy}
        \label{fig:hierarchy}
    \end{subfigure}
    \caption{\textbf{Depiction of the 2D embeddings of the STL-10 validation dataset.} The learnt embeddings of our proposed hyperbolic MSN with ideal prototypes. The red points represent the prototypes, the dotted line is the boundary of the Poincar\'e ball. For (a) natural semantic class clusters form at individual prototypes. Neighbouring prototypes capture similar semantic class sub-features (b), this is observed clearly with fire trucks being separately clustered to trucks (\textit{purple points}), and large grilled cars (\textit{light-red points}) being positioned in a similar manner. Seaplanes are positioned alongside boats (\textit{blue points}) rather than airplanes (\textit{dark red}) yet lie closer to the origin given there ambiguity.}
    \label{fig:2d_vis}
\end{figure}
To summarise, the main contributions of the paper are the following:
\begin{itemize}
    \setlength\itemsep{0.1em}
    \item We propose a hyperbolic reformulation of the MSN clustering-based loss function, Hyperbolic Masked Siamese Networks (HMSN).
    \item We utilise ideal prototypes that lie on the ideal boundary of the Poincar\'e ball to encourage full utilisation of the space. We introduce Hyperbolic Masked Siamese Networks with Ideal Prototypes (HMSN-IP), based on the MSN method, employing the Busemann prototype loss from metric learning as a measure of distance between embeddings and the ideal prototypes. 
    \item We propose to project Euclidean representations to the Poincar\'e ball of Hyperbolic space at the output of the encoder and present the use of hyperbolic projection heads as a solution to preserve hyperbolic structure in the output of the Euclidean encoder for downstream tasks.
    \item We empirically demonstrate that both our propositions outperform the Euclidean counterparts on few-shot and low-shot learning tasks in fewer embedding dimensions, whilst remaining competitive in linear evaluation tasks.
\end{itemize}

\section{Related Work}
\paragraph{Self-Supervised Representation Learning.}
In self-supervised learning, we consider a set of unlabelled images $D$ which we aim to learn representations for use in downstream tasks. We pre-train on $D$ and then adapt the representations via a supervised task using a set of images $S$ and their corresponding labels where here $\|S\| << \|U\|$. The most successful methods to learn good representations employ view-invariant joint-embedding architectures \citep{oord2018representation,wu2018unsupervised,hjelm2018learning,bachman2019learning,he2020momentum,misra2020self,bardes2021vicreg, assran2023self} which aim to predict the embedding of a view from another view of the same image. There exist a number of methods to train joint embedding predictive architectures, non-contrastive, which maximise the information content of the embeddings \citep{grill2020bootstrap, chen2020simple, bardes2021vicreg}, and distillation, in which the outputs of one branch of the Siamese join embedding architecture act as a target for the other branch \citep{caron2018deep,caron2019unsupervised,zhuang2019local,caron2020unsupervised,yan2020clusterfit,asano2019self}. The latter is the focus of this work, primarily the methodologies DINO \citep{caron2021emerging} and its later derivation MSN \citep{assran2022masked}, which utilise discrete cluster prototypes to quantise the output representations. 

Clustering approaches have excelled with the use of vision transformers achieving near-to or state-of-the-art performance in most self-supervised benchmarks \citep{caron2020unsupervised, caron2021emerging, assran2022masked}. More recently there has been greater thought placed into masking strategies of these approaches with the aim to learn better representations through prediction or invariance to the missing regions \citep{zhou2021ibot, xie2022simmim, assran2022masked, he2022masked, chang2023muse}. These approaches are particularly of interest in this work as a result of their exceptional performance in low-shot and extreme low-shot training settings \citep{assran2022masked}. As such, given our aims to improve low-shot learning, we base our loss and hyperbolic representation space reformulations on the leading architectural designs, specifically MSN.

\paragraph{Hyperbolic Learning.}
The advocation for learning representations or embeddings in non-Euclidean space in deep learning has, in recent years, increased rapidly. Hyperbolic reformulations of deep learning layers across both intermediate \citep{ganea2018hyperbolic} and classification layers \citep{ghadimiatigh2022hyperbolic, khrulkov2020hyperbolic}, as well as whole architectural propositions \citep{shimizu2020hyperbolic} have been proposed with improved performance and computational efficiency. Hyperbolic deep learning has seen great success in tasks where the representation of tree-like structures is beneficial like natural language \citep{aly2019every,tifrea2018poincar,zhang2021hyperbolic} and graph neural networks \citep{liu2019hyperbolic,chami2019hyperbolic,bachmann2020constant}. The application of hyperbolic deep learning in vision is still however foundational, yet vast work has been undertaken in visual metric learning \citep{bi2015beyond,khrulkov2020hyperbolic,ermolov2022hyperbolic}, with \citep{yan2021unsupervised} performing hierarchical unsupervised similarity based metric learning, \citep{ermolov2022hyperbolic} extending the DINO \citep{caron2021emerging} architecture with hyperbolic contrastive learning for metric learning. The latter which projects the output embedding space to hyperbolic space further motivates our decision to base our work on MSN given its known capabilities, albeit not in a self-supervised setting.

Moreover, hyperbolic metric learning and prototype learning approaches have demonstrated their capabilities in few-shot and zero-shot \citep{fang2021kernel, xu2022meta, ghadimi2021hyperbolic} learning tasks, outperforming Euclidean embedding methods by some margins. Given the connections between metric learning, and prototype learning to self-supervision, there exhibit clear enablers between the domains. The work \citep{ermolov2022hyperbolic} explores these, initially investigating hyperbolic self-supervised learning before re-evaluating it as a metric learning approach given improved performance in this domain. Contrastive self-supervision has also been addressed in \citep{ge2022hyperbolic, yue2023hyperbolic} which proposes a number of hyperbolic reformulations of prominent SSL and contrastive objectives. Our work aims to further explore the use of hyperbolic embedding space for self-supervised learning, advocating for its use to help provide greater insights and representation quality for all tasks while leveraging its strong performance in few-shot and low-shot learning.

\section{Prerequisites}
\subsection{Hyperbolic Learning: The Poincar\'e Ball Model}
Hyperbolic space $\mathbb{D}^d$ is the unique simply connected $d$-dimensional Riemannian manifold of constant negative curvature, where curvatures measure the deviation from flat Euclidean geometry. The constant negative curvature of the hyperbolic space, although analogous to the Euclidean sphere, presents some significant differences in geometric properties. As such hyperbolic space cannot be isometrically embedded into Euclidean space, yet there exist a number of conformal models of hyperbolic geometry \citep{cannon1997hyperbolic} employing hyperbolic metrics providing a subset of Euclidean space. In this work, we employ the Poincar\'e ball model for hyperbolic geometry given its wide adoption in computer vision and unique properties ideal for embedding between euclidean and hyperbolic representations. The Poincar\'e ball model $(\mathbb{D}_{c}^{d}, g^{\mathbb{D}_c})$ is defined by the manifold $\mathbb{D}_{c}^{d}=\{x\in \mathbb{R}^d : c \|x\|^2 <1\}$ with the Riemannian metric 

\begin{equation}
    g^{\mathbb{D}_c} = (\lambda^c_x)^2 g^E = \left(\frac{2}{1-c\|x\|^2}\right)^2 \mathbb{I}^d
\end{equation}
where $g^E=\mathbb{I}^d$ is the Euclidean metric tensor and $\lambda^c_x=\frac{2}{1-c\|x\|^2}$ is the conformal factor with $c$, a hyperparameter, controlling the curvature and radius of the ball. The conformal factor scales the local distances which approach infinity near the boundary of the ball, providing the unique property of space expansion. Such space expansion of hyperbolic spaces makes them continuous analogues of trees, given volumes of an object with diameter $r$ scale exponentially with $r$. Thus, when referring to a tree with branching factor $k$, there are $\mathcal{O}(k^l)$ nodes at level $l$, where $l$ serves as a discrete analogue of the radius. This is the fundamental property which the advocating work \citep{ganea2018hyperbolic, khrulkov2020hyperbolic, yan2021unsupervised} and ours takes advantage of, allowing for the efficient embedding of natural hierarchies \citep{sarkar2011low}. 

Our approach employs encoders that operate in Euclidean space, and as such, we need to define a bijection from Euclidean embeddings of the encoder to the Poincar\'e ball of hyperbolic space. To achieve this we apply an exponential map $\textnormal{exp}^c_{v}(x): \mathbb{R}^d \rightarrow \mathbb{D}_{c}^{d}$ on Euclidean vector $x$ with some fixed base point $v \in \mathbb{D}_{c}^{d}$ which we set $v$ to be the origin, simplifying the exponential map and measures of distance which will be defined later. The exponential map is as follows,
\begin{equation}
    \textnormal{exp}^c_{v}(x) = v{\oplus}_{c} \left( \textnormal{tanh} \left( \sqrt{c}\frac{\lambda^c_v \|x\|}{2} \right) \frac{x}{\sqrt{c}\|x\|} \right)
    \label{eq:exp-map}
\end{equation}
with its inverse logarithm map given by
\begin{equation}
    \textnormal{log}^c_{v}(x) = \frac{2}{\sqrt{c}\lambda^c_v}\textnormal{arctanh} \left( \sqrt{c}\|-v{\oplus}_{c}x\|\right)\frac{-v{\oplus}_{c}x}{\|-v{\oplus}_{c}x\|} .
    \label{eq:log-map}
\end{equation}
Given the change in geometry, hyperbolic spaces do not allow for standard vector space operations, as such we employ gyrovector formalism for standard operations such as addition, subtraction, multiplication \citep{ungar2008gyrovector, ganea2018hyperbolic}. Therefore, from Eq. \ref{eq:exp-map}, ${\oplus}_{c}$ is defined as the gyrovector or M\"obius addition of a pair of points $x,y \in \mathbb{D}_{c}^{d}$
\begin{equation}
    v{\oplus}_{c}w =\frac{(1+2c\langle v,w \rangle +c\|w\|^2)v+(1-c\|v\|^2)w}{1+2c\langle v,w \rangle + c^2\|v\|^2\|w\|^2} .
\end{equation}
Leading from gyrovector formalism is the notion of distance, vital for self-supervised losses where typically the Euclidean cosine similarity and distance, are employed \citep{richemond2020byol,chen2020simple,bardes2021vicreg}. On the Poincar\'e ball of hyperbolic space we define the distance between $x,y \in \mathbb{D}_{c}^{d}$ as follows:
\begin{equation}
    \text{dist}_{\mathbb{D}}(x,y) = \frac{2}{\sqrt{c}}\textnormal{arctanh}\left( \sqrt{c}\|-x\oplus_c y\|\right),
    \label{eq:dist}
\end{equation}
which with $c=1$ the geodesic is recovered, a vital concept given cosine similarity is analogous to sphere geodesic distance, whereas $c \rightarrow 0$ the Euclidean distance is produced. 

\subsection{Self-Supervised Learning: Masked Siamese Networks}
In this work, we use the Masked Siamese Network (MSN) \citep{assran2022masked} as a base for our hyperbolic implementation due to its leading performance as a few-shot learner in self-supervision, its computational efficiency, and established clustering based loss formulation. This therefore provides us with the best opportunity for baseline comparison when striving for improved low-shot performance when employing hyperbolic representation space. 

In MSN data augmentations are applied to image $\textbf{x}_i$ to produce the target view $\textbf{x}^{+}_i$ and a set of $M \geq 1$ anchor views $\textbf{x}_{i,1}, \textbf{x}_{i,2}, \dotsc, \textbf{x}_{i,M}$ where $i$ is the index of the sample mini-batch of images $B \geq 1$. The anchor views $\textbf{x}_{i,m}$ are subsequently patched into $N \times N$ non-overlapping regions and masked randomly or via a focal scheme \cite{assran2022masked} denoted by $\hat{\textbf{x}}_{i,m}$. The encoders $f_{\theta}$ and $f_{\Bar{\theta}}$ are identical although differently parameterised trunks of the ViT \cite{dosovitskiy2020image} outputting representations corresponding to the [CLS] token. The anchor views $\hat{\textbf{x}}_{i,m}$ processed by the anchor encoder which is parameterised by $\theta$ produce the representations $z_{i,m} \in \mathbb{R}^d$ while the target views $\textbf{x}^{+}_i$ processed by the target encoder parameterised by $\Bar{\theta}$ produce the representations $z^+_{i} \in \mathbb{R}^d$. The target encoder is not directly updated by the optimisation process with gradients only computed with respect pot the anchor predictions, rather $\Bar{\theta}$ are updated via an exponential moving average of the anchor encoder. Each encoder is trained with a 3-layer non-linear projection head $g_\theta(\cdot)$ and $g_{\Bar{\theta}}(\cdot)$ with batch-normalisation at the input and hidden layers, which is later discarded during evaluation.

The metric which drives invariance between views is the soft distribution over a set of $K > 1$ learnable prototypes of dimension $d$ denoted by $\textbf{q} \in \mathbb{R}^{K \times d}$. The distribution is computed as the cosine similarity between the prototypes $\textbf{q}$ and the $L_2$-normalized anchor and target views pairs where for the anchor view representation $z_{i,m}$ the prediction distribution $p_{i,m} \in \Delta_K$ is given by 

\begin{equation}
    p_{i,m} := \text{softmax} \left( \frac{z_{i,m} \cdot \textbf{q}}{\tau} \right).
    \label{eq:euc-sim}
\end{equation}

The same formulation applies to the target view representations $z^+_{i}$ substituting the anchor views to produce target predictions $p^+_{i} \in \Delta_K$. The temperature $\tau \in (0,1)$ is always chosen to be larger for the anchor predictions ($\tau^+ < \tau$) to encourage sharper target predictions producing confident low entropy anchor predictions which have been shown to provably discourage collapsing solutions \cite{assran2022masked}. 

The network is trained by the cross-entropy loss $H(p^+_i, p_{i,m})$ to penalise differing predictions of views that originate from the same image. This cross-entropy loss is regularised by mean entropy maximisation to encourage the use of the full set of prototypes, which maximises the entropy of the mean anchor predictions $H(\bar{p})$. The overall objective to be minimised when optimising over $\theta$ and $q$ is given by Eq.\ref{eq:euc-obj} where $\lambda$ controls the weight of the mean entropy maximisation regularisation.

\begin{equation}
    \frac{1}{MB}\sum^{B}_{i=1}\sum^{M}_{m=1}H(p^{+}_{i}, p_{i,m}) - \lambda H(\bar{p})
    \label{eq:euc-obj}
\end{equation}

For a more detailed description of MSN, we refer the reader to \citep{assran2022masked}, and for implementation details we refer to the supplementary material.

\section{Hyperbolic Masked Siamese Networks}\label{sec:hmsn}

 Learning of hyperbolic embeddings under the MSN framework can most simply be achieved by mapping the euclidean output embeddings of the network to the Poincar\'e ball model, via Eq.\ref{eq:exp-map} followed by the substitution of the euclidean vector operations of the objective function Eq.\ref{eq:euc-obj} for hyperbolic gyrovector equivalents. This approach to hyperbolic reformulation has shown to be an effective method of learning hyperbolic visual representations \citep{ermolov2022hyperbolic} and in contrastive self-supervision \citep{ge2022hyperbolic, yue2023hyperbolic}. We therefore first follow this methodology to examine the capabilities of hyperbolic self-supervision, we refer to the reformulation as Hyperbolic Masked Siamese Network (HMSN). We begin by projecting the anchor and target representations to the Poincar\'e ball model by the exponential map Eq.\ref{eq:exp-map}, and initialising the prototypes $\textbf{q}$ normally on the same hyperbolic space. The standard euclidean cosine similarity in Eq. \ref{eq:euc-sim} to compute the prediction metric $p_{i,m}$, is substituted for the geodesic distance of Eq.\ref{eq:dist}. The reformulation of Eq. \ref{eq:euc-sim} results in the following prediction:

\begin{equation}
    p^{\mathbb{D}}_{i,m} := \text{softmax} \left( \frac{ \text{dist}_{\mathbb{D}}(z_{i,m} , \textbf{q})}{\tau} \right).
\end{equation}

\begin{figure}[h!]
\centering
\begin{minipage}{.5\textwidth}
  \centering
  \includegraphics[width=0.8\linewidth]{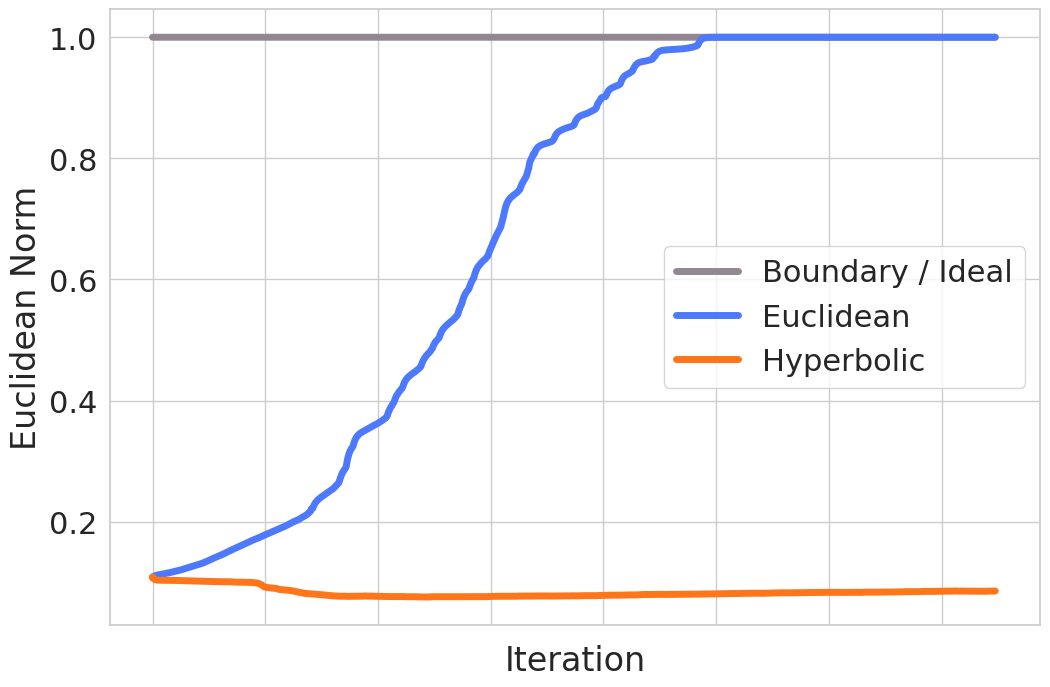}
  \captionof{figure}{\textbf{Distance of prototypes from origin during training.} Mean Euclidean norm of Prototypes during training of Euclidean MSN baseline (\textit{blue}), Hyperbolic MSN (\textit{orange}), and Hyperbolic MSN with ideal prototypes (\textit{grey}).}
  \label{fig:euclid_dist}
\end{minipage}\hspace{.55em}
\begin{minipage}{.44\textwidth}
  \centering
    \begin{subfigure}[b]{0.5\textwidth}
         \centering
         \includegraphics[width=0.55\textwidth]{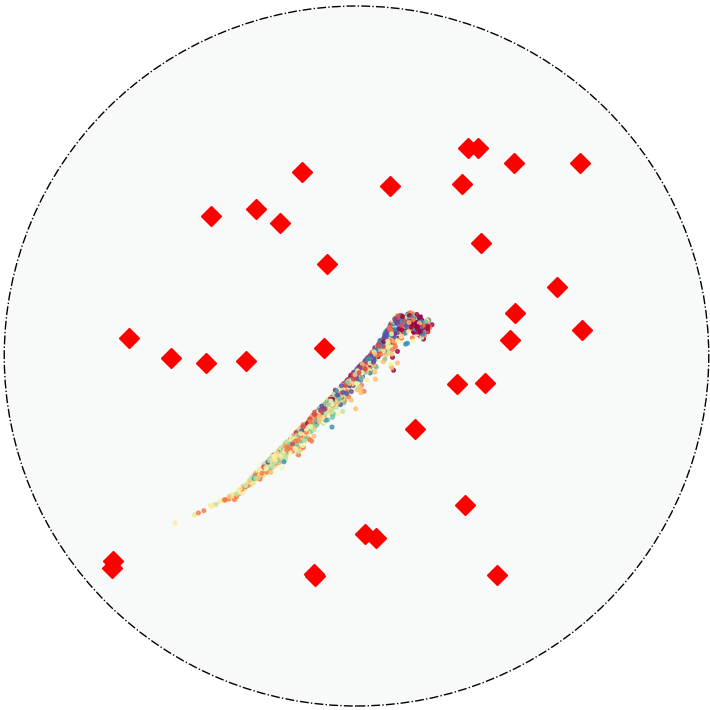}
         \caption{1 Epochs}
         \label{fig:HMSN_1}
     \end{subfigure}%
     \begin{subfigure}[b]{0.5\textwidth}
         \centering
         \includegraphics[width=0.55\textwidth]{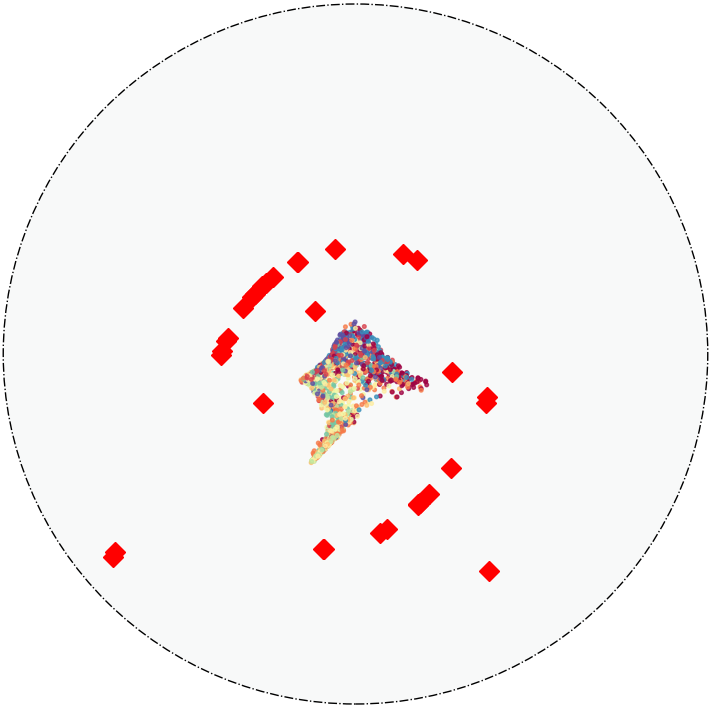}
         \caption{5 Epochs}
         \label{fig:HMSN_5}
     \end{subfigure}

     \begin{subfigure}[b]{0.5\textwidth}
         \centering
         \includegraphics[width=0.55\textwidth]{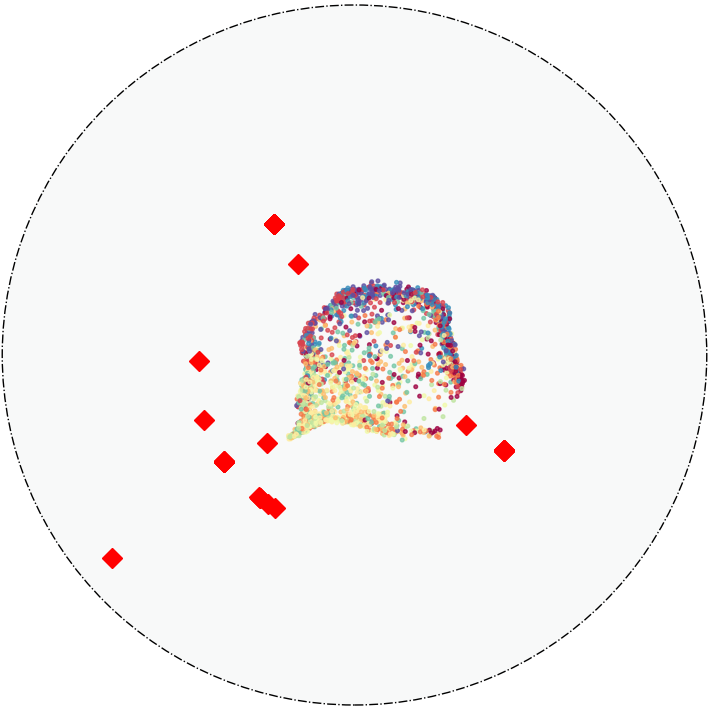}
         \caption{10 Epochs}
         \label{fig:HMSN_10}
     \end{subfigure}%
     \begin{subfigure}[b]{0.5\textwidth}
         \centering
         \includegraphics[width=0.55\textwidth]{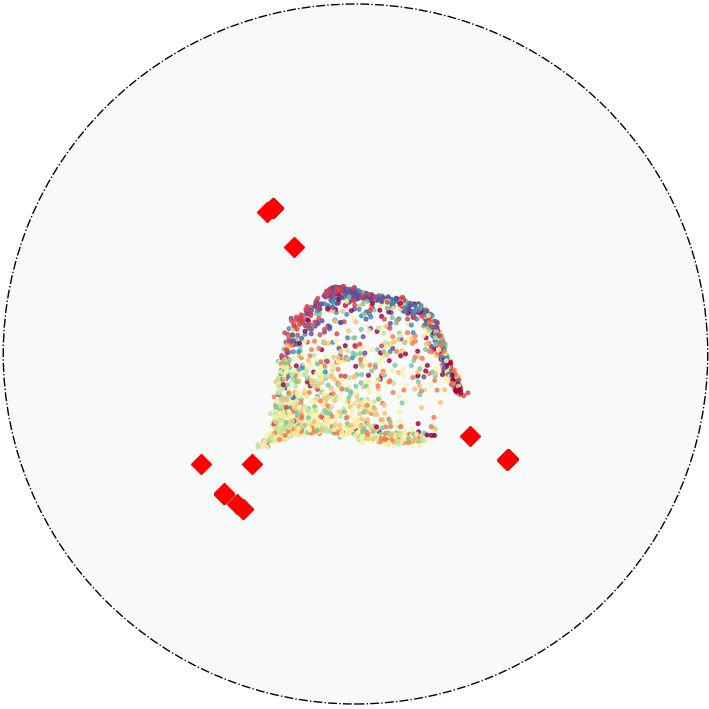}
         \caption{20 Epochs}
         \label{fig:HMSN_20}
     \end{subfigure}

  \captionof{figure}{\textbf{2D visualisation of learnable prototype positioning.} During early stages of HMSN training.}
  \label{fig:proto_shrink}
\end{minipage}
\end{figure}

The overall objective function in Eq.\ref{eq:euc-obj} remains identical substituting $p_{i,m}$ for $p^{\mathbb{D}}_{i,m}$ although is trained by Riemannian Adam \citep{becigneul2018riemannian} as we are directly optimising jointly the prototypes in hyperbolic space and the euclidean parameters $\theta$. We initialise the prototypes to be normal with a small standard deviation ($0.01$) centred around the origin for improved stability early in training. We also clip the Euclidean representations before projection to the Poincar\'e ball model as in \citep{guo2022clipped} to assist in vanishing gradients when backpropagating from the hyperbolic space to the euclidean space as embeddings tend towards the boundary of hyperbolic space during training.

\section{Hyperbolic Masked Siamese Networks with Ideal Prototypes}\label{sec:hybip}
Learning prototypes by HMSN exhibits behaviour that results in the under utilisation of the hyperbolic embedding space. During training, the learnable prototypes tend towards the origin and converge to a region that is significantly distant from the boundary of the space, this phenomena is visualised in Figure \ref{fig:euclid_dist} and Figure \ref{fig:proto_shrink}. This in-turn restricts the embedding space resulting in uncertain and less abstract positioning. A solution is to encourage the prototypes to lie closer to the boundary through additional regularisation term maximising euclidean norm, or to provide some prior to the embeddings to place them more akin to the hierarchies we aim to capture \citep{ghadimiatigh2022hyperbolic}. However, The former is naive and the latter requires annotations from human observers, not feasible in the self-supervised setting. To address this we place the prototypes at ideal points of the Poincar\'e ball. 

The ideal points, $\mathbb{I}_d$, are positioned prior to training based on separation on the unit hypersphere $\mathbb{S}_d$ for $d \ge 3$ while positioned uniformly on $\mathbb{S}_d$ when $d = 2$ given ideal points of the hyperbolic space $\mathbb{D}_d$ are homeomorphic to $\mathbb{S}_d$ \citep{ghadimiatigh2022hyperbolic}. As the set of ideal points lies on the boundary of the hyperbolic space, the geodesic distance Eq.\ref{eq:dist} from an ideal point to any point in hyperbolic space is infinite. Therefore, to measure the assignment of a hyperbolic embedding to an ideal prototype the Busemann loss function is used. In the Poincar\'e ball model, the Busemann function is given by Eq.\ref{eq:buse}.

\begin{equation}
    b_{\mathbf{q}}(z_{i,m}) = \log \frac{\|\mathbf{q} - z_{i,m}\|^2}{(1-\| z_{i,m} \|^2)}
    \label{eq:buse}
\end{equation}

The Busemann function \citep{busemann2012geometry} is considered to be a distance measured to infinity defined in any space. As with the hyperbolic reformulation in Section \ref{sec:hmsn} we can substitute the cosine similarity for the Busemann function and position the prototypes at ideal points to produce the following prediction.  

\begin{equation}
    p_{i,m}^{\mathbb{I}} := \text{softmax} \left( \frac{-b_{\mathbf{q}}(z_{i,m})}{\tau} \right).
    \label{eq:hyb-ip}
\end{equation}

An important distinction from the work in \cite{ghadimi2021hyperbolic} is that our function does not require a penalty term to penalise the overconfidence of the embeddings. Instead, the temperature $\tau$ scaling of the Softmax in Eq. \ref{eq:hyb-ip} increases the magnitude of the hyperbolic embedding as $\tau$ decreases \cite{guo2022clipped}. As a result, the embeddings are prevented from approaching the boundary of the ball as the Softmax is sharpened and certainty increases. In practice, tuning $\tau$ for performance whilst ensuring the embeddings do not lie on the boundary -- the cause of vanishing gradients -- is non-trivial. Instead, we clip the euclidean representations before the exponential mapping as done in \ref{sec:hmsn}. To avoid collapsed representations we introduce an entropy term to encourage unique prototype assignment \citep{assran2022masked}, the resulting objective is,

\begin{equation}
    \frac{1}{MB}\sum^{B}_{i=1}\sum^{M}_{m=1}H(p^{\mathbb{I}+}_{i}, p_{i,m}^{\mathbb{I}}) - \lambda H(\bar{p}^{\mathbb{I}}) + \beta H(p_{i,m}^{\mathbb{I}}).
    \label{eq:hyb-obj}
\end{equation}

\begin{figure}[t]
    \centering
    \begin{subfigure}[b]{0.54\textwidth}
         \centering
         \includegraphics[width=1.0\textwidth]{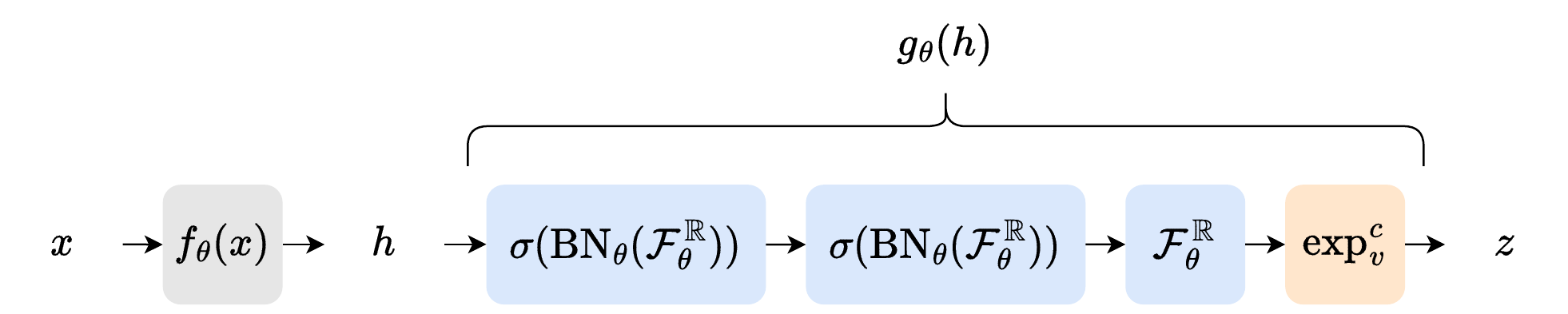}
            \caption{}
         \label{fig:std_proj}
    \end{subfigure}
    \hfill
    \begin{subfigure}[b]{0.45\textwidth}
         \centering
         \includegraphics[width=1.0\textwidth]{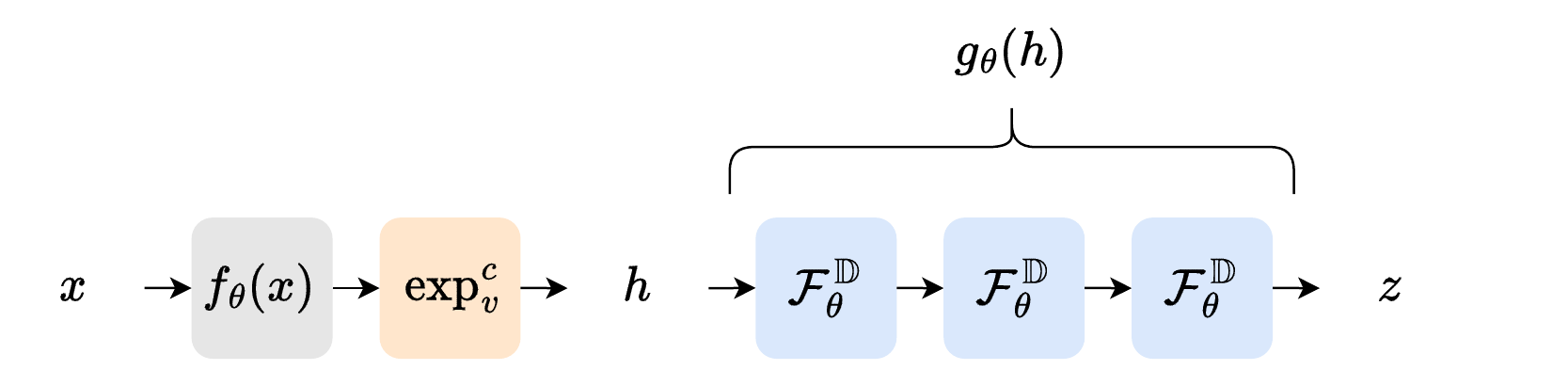}
        \caption{}
         \label{fig:hyb_proj}
    \end{subfigure}
    \caption{\textbf{Visualisation of the projection head architectures:} (a) The Euclidean projection head projects the euclidean embeddings to the Poincar\'e ball before the computation of the loss. (b) The Hyperbolic projector receiving Poincar\'e ball hyperbolic representations from the encoder.}
    \label{fig:proj}
\end{figure}
\vspace{-2em}
\section{Hyperbolic Projection Head}
Most notable approaches to hyperbolic self-supervised learning propose solutions that are akin to the reformulation procedure aforementioned \citep{ermolov2022hyperbolic, ge2022hyperbolic, yue2023hyperbolic}. However, all such methods fail to address that the representations used for downstream tasks remain euclidean. We instead project the representations to hyperbolic space at the output of the encoder trunk before the projection head, $g_\theta(\cdot)$, rather than at the output of the projection head as done by prior methods (Figure \ref{fig:proj}). The motivation is that the projection head is present during training and then removed for downstream tasks which is when we intend to most effectively utilise the structure captured in hyperbolic space. Recent works \citep{bordes2022guillotine} which examine the role of the projection head have identified its role to remove the training tasks overfitting bias, which when removed results in representations that are significantly more generalisable to downstream tasks.

Given that we propose to project to the hyperbolic space before the projection head, the Multi-Layer Perception (MLP) comprised of three fully-connected layers and their non-linear activation must remain hyperbolic to preserve the embedding structure. To achieve this, we employ a Hyperbolic projection head comprised of three Hyperbolic fully-connected layers, $\mathcal{F}_\theta^{\mathbb{D}}$, define in \cite{shimizu2020hyperbolic}(Figure \ref{fig:hyb_proj}) followed by a hyperbolic ReLU non-linearity \citep{van2023poincar} with exception to the final layer.
We examine in more detail the effect of the projection head in Section \ref{sec:proj}.

\section{Experimentation}
To examine the quality of hyperbolic representations, a number of standardised benchmark tasks are performed. We first evaluate the representations learnt by the ViT encoder on the ImageNet-1K dataset under linear evaluation, followed by few-shot evaluation on ImageNet-1K using only 1\% of the labelled training images per class\footnote{ImageNet subsets: \href{https://github.com/facebookresearch/msn}{https://github.com/facebookresearch/msn}} as per \citep{assran2022masked}. Given our representations lie in hyperbolic space we cannot directly compare using Euclidean classifiers, as such, we employ a hyperbolic multi-linear regression classifier with implementation identical to that reported in \citep{shimizu2020hyperbolic}. For all methods, we pre-train with a batch size of 1024 \footnote{ViT-S/16 Trained on 6x Nvidia A100 80GB GPUs for 800 epochs $\approx 150$ hours.}, producing views identically to \citep{assran2022masked} with 1 anchor, 1 random mask, and 10 focal mask views. 

\begin{table}[h!]
    \centering
    \caption{\textbf{Linear Classification on ImageNet-1K}. Top-1 accuracy for linear models trained on frozen
features from different self-supervised methods.}
    \begin{tabular}{l|llllc}
         \textbf{Method} & \textbf{Arch.} & \textbf{Params.} & \textbf{Epochs} & \textbf{Dims.} & \textbf{Top-1 (\%)} \\
         \midrule
         SimCLR v2 \citep{chen2020improved} & RN50 & 24M & 800 & 2048 & 71.7\\
         BYOL \citep{grill2020bootstrap} & RN50 & 24M & 1000 & 2048 & 74.4\\
         Barlow-T \citep{zbontar2021barlow} & RN50 & 24M & 1000 & 8192 & 73.2\\
         VICReg \citep{bardes2021vicreg} & RN50 & 24M & 1000 & 8192 & 73.2\\
         DINO \citep{caron2021emerging} & ViT-S/16 & 22M & 800 & 2048 & 77.0\\
         iBOT \citep{zhou2021ibot} & ViT-S/16 & 22M & 800 & 8192 & 77.9\\
         MSN \citep{assran2022masked} & ViT-S/16 & 22M & 800 & 256 & 76.9\\
         \midrule
         \rowcolor{LightCyan} HMSN (\textit{ours}) & ViT-S/16 & 22M & 800 & 64 & 76.0\\
         \rowcolor{LightCyan} HMSN-IP (\textit{ours}) & ViT-S/16 & 22M & 800 & 64 & 76.8\\
        
    \end{tabular}
    \label{tab:lin-eval}
\end{table}

\subsection{Linear Evaluation}
We train a hyperbolic linear classifier on the labelled ImageNet-1K training set on the representations produced by of our frozen pre-trained, self-supervised hyperbolic vision transformer. Table \ref{tab:lin-eval} reports the top-1 linear evaluation accuracies (\%) of both our proposed method compared against other leading approaches on the ImageNet-1K validation set, the results are the average of 3 randomly initialised runs. The hyperbolic reformulation (HMSN) performs marginally worse that its euclidean baseline albeit with fewer embedding dimensions 64 instead of 256. On the other hand, the Hyperbolic Masked Siamese Networks with Ideal Prototypes (HMSN-IP) performs comparatively to the MSN baseline showing a 0.1\% difference with the same reduction in embedding dimensions. Encouraging, a performance drop is not observed in HMSN-IP given the fixed ideal prototypes. We note that training HMSN results in uniformly distributed prototypes akin to the ideal prototypes, albeit positions far closer to the origin restricting the representation space (Figure \ref{fig:euclid_dist}).
%
\begin{table}[h!]
    \centering
    \caption{\textbf{Low-shot Linear Evaluation on ImageNet-1K}. Top-1 Accuracy for linear models trained on frozen features from different methods, fine-tuning only uses 1\% of the labels.
}
    \begin{tabular}{llllc}
         \textbf{Method} & \textbf{Arch.} & \textbf{Params.} & \textbf{Dims.} & \textbf{Top-1 (\%)} \\
         \midrule
         Barlow-Twins \citep{zbontar2021barlow} & RN50 & 24M & 8192 & 55.0\\
         SimCLR v2 \citep{chen2020improved} & RN50 & 24M & 2048 & 57.9\\
         PAWS \citep{assran2021semi} & RN50 & 24M & 2048 & 66.5\\
         DINO \citep{caron2021emerging} & ViT-S/16 & 22M & 2048 & 64.5\\
         iBOT \citep{zhou2021ibot} & ViT-S/16 & 22M & 8192 & 65.9\\
         MSN \citep{assran2022masked} & ViT-S/16 & 22M & 256 & 67.2\\
         \midrule
         \rowcolor{LightCyan} HMSN (\textit{ours}) & ViT-S/16 & 22M & 64 & 67.6\\
         \rowcolor{LightCyan} HMSN-IP (\textit{ours}) & ViT-S/16 & 22M & 64 & 68.7\\
        
    \end{tabular}
    \label{tab:low-shot}
\end{table}

\subsection{Low-Shot Linear Evaluation} \label{sec:low-shot}
The ability to learn representations from unlabelled data that are of high enough quality to be used in downstream tasks with very few labelled examples is the key motivator behind self-supervised learning. Moreover, our design decisions to employ hyperbolic representation space to learn hierarchies and as such represent semantic concepts in a more structured manner are driven by the goal of improving few-shot learning. As with the linear evaluation, we pre-train our encoder on the ImageNet-1K dataset, freezing the weights and training a linear classifier on top using a subset of the ImageNet-1K labelled training set. The performance of our self-supervised models by performing linear evaluation on very few labelled examples for each class is reported in Table \ref{tab:low-shot}. 

In the standard low-shot benchmark 1\% of the ImageNet-1K labels are employed for linear evaluation (approximately 13 images per class), the results are presented in Table \ref{tab:low-shot} alongside alternative competitive self-supervised methods. Our hyperbolic reformulation (HMSN) outperforms its Euclidean counterpart with a 0.4\% performance improvement with the ViT-S/16. The extension, Hyperbolic Masked Siamese Networks with Ideal Prototypes (HMSN-IP) sees a further 1.0\% improvement over the hyperbolic reformulation, HMSN. 

\begin{wraptable}[10]{R}{7cm}
    \centering
    \captionof{table}{\textbf{Hyperbolic and Euclidean Projection Heads.} Linear evaluation accuracy on the Imagenet-1K validation set training for both Hyperbolic ($\mathbb{D}$) and Euclidean ($\mathbb{R}$) classifiers.}
    \begin{tabular}{ccc}
        \textbf{Projector $g(\cdot)$} & \textbf{ $\mathbb{R}$ Top-1 (\%)} & \textbf{ $\mathbb{D}$ Top-1 (\%)}\\
        \midrule
        Euclidean & 58.1 & 52.0\\
        Hyperbolic & 48.1 & 66.2
    \end{tabular}
    \label{tab:mlp_proj}
\end{wraptable}
\subsection{Projection Head}\label{sec:proj}
The approaches described in \ref{sec:hmsn} and \ref{sec:hybip} both make the important distinction from previous works \citep{ge2022hyperbolic, yue2023hyperbolic} regarding the projection head in the training procedure. Typically the projection head is disregarded from the reformulation of euclidean SSL methods into hyperbolic ones, where embeddings are typically hyperbolic and representations remain Euclidean for comparison in downstream tasks (visually depicted in \ref{fig:proj}). Therefore, hyperbolic properties are lost when utilising the representations in downstream tasks.

Table \ref{tab:mlp_proj} reports the linear evaluation top-1 accuracy on the ImageNet-1K validation set for a pre-trained ViT-S/16 by HMSN-IP loss for 100 epochs \footnote{Best performance is not achieved by 100 epochs, however, the results allow for reasonable ablations.} with either a Euclidean or Hyperbolic projection head. The downstream linear evaluation top-1 accuracy are given for both a Euclidean linear evaluation procedure as described in \citep{assran2022masked} and the Hyperbolic linear evaluation procedure (further details given in Supplementary Material). The results clearly demonstrate that when evaluating using a downstream hyperbolic classifier the HMSN-IP with the hyperbolic projection head produces representations that are of a higher hyperbolicity compared to Euclidean counterpart. 
%
%
\begin{wrapfigure}[18]{R}{6cm}
    \centering
    \vspace{-1.2em}
    \includegraphics[width=0.38\textwidth]{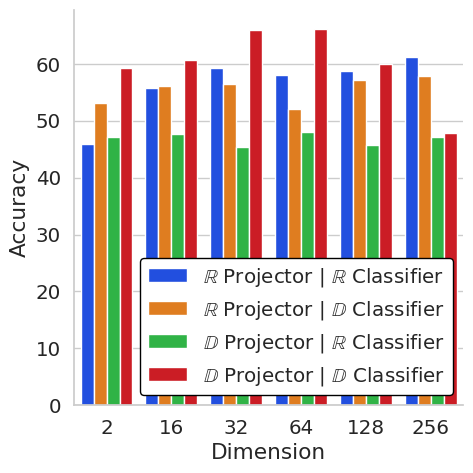}
    \captionof{figure}{\textbf{Embedding Dimensions.} Linear evaluation accuracy on the Imagenet validation set training for Hyperbolic ($\mathbb{D}$) and Euclidean ($\mathbb{R}$) classifiers and projectors.}
        \label{fig:dims}
\end{wrapfigure}
\subsection{Embedding Dimensions}\label{sec:dims}
An important and unique property of the negative curvature hyperbolic space is the exponentially expanding volume with respect to distance from the origin. This results in a representation space that exhibits the volume necessary for separability at far fewer dimensions. To access this, we pre-train ViT-S/16 with baseline MSN and HMSN-IP on ImageNet-1K for 100 epochs under different output dimensions.

We report in Figure \ref{fig:dims} the linear evaluation top-1 accuracy on the ImageNet-1K test of both Euclidean and Hyperbolic classifiers for each given dimension and projection head. We can see the expected increase in performance when the Euclidean projector dimensions are increased (\textit{blue} and \textit{orange} bars), this expected result aligns with previous investigations of the projection head \citep{bordes2022guillotine}. For the Hyperbolic projector and hyperbolic classifier setting (\textit{red}), we observe steady increase in accuracy until a significant drop-off occurred at 128 dimensions. 
\section{Conclusion}
This work investigates the hyperbolic self-supervised learning, introducing a hyperbolic extension to the Masked Siamese Network model where we empirically show improved downstream performance in hyperbolic classifiers for linear evaluation, transfer learning, and low-shot learning. We further improve on this by introducing a method that instead uses prototypes placed on the ideal boundary of the Poincar\'e ball model. We empirically demonstrate that this method improves low-shot downstream task performance over the standard hyperbolic reformulation. Both our proposed methods outperform or perform competitively to their Euclidean counterparts but do so at fewer embedding dimensions (Figure \ref{fig:dims}) whilst exhibiting clear semantic class hierarchies (Figure \ref{fig:hierarchy}). 

\paragraph{Limitations \& Broader Impact}
Our work aims to produce better representations of images in setting where data annotations are scarce, it can therefore be seen how such methods can lead to more accurate or informative models for a number of downstream tasks with positive societal impact. However, as is the case with all vision systems, there is potential for exploitation and security concerns and one should take into consideration AI misuse when extending our method.

Hyperbolic self-supervision can improve the compactness of these representations, therefore providing promising research directions in applications such as data transmission and compression via SSL. Importantly, the improved interpretability due to uncertainty proxy of learned representations by embedding latent tree-like hierarchies leads to exciting new SSL understanding. However, computing in the hyperbolic space introduces challenges regarding matrix and vector operations and as such, there exists implementation and computational difficulties compared to Euclidean approaches. In practice, we do not find these significantly impactful at the presented scale, regardless, future work or extensions should take care in their case. 


\bibliography{ref}
\bibliographystyle{abbrv}

\end{document}